\documentclass[conference]{IEEEtran}
\IEEEoverridecommandlockouts
 
\usepackage{cite}
\usepackage{amsmath,amssymb,amsfonts}
\usepackage{algorithmic}
\usepackage[ruled,norelsize]{algorithm2e}
\usepackage{graphicx}
\usepackage{textcomp}
\usepackage{xcolor}
\usepackage{hyperref}
\def\BibTeX{{\rm B\kern-.05em{\sc i\kern-.025em b}\kern-.08em
    T\kern-.1667em\lower.7ex\hbox{E}\kern-.125emX}}
    
\begin{document}

\title{LAIF: AI, Deep Learning for Germany Suetterlin Letter Recognition and Generation
}

\author{\IEEEauthorblockN{Enkhtogtokh Togootogtokh} 
\IEEEauthorblockA{\textit{Technidoo Solutions Lab} \\
\textit{Technidoo Solutions Germany and Mongolian University of Science and Technology}\\
Bavaria, Germany \\
enkhtogtokh.java@gmail.com, togootogtokh@technidoo.com} 
\and 
\IEEEauthorblockN{Christian Klasen}
\IEEEauthorblockA{\textit{Technidoo Solutions Lab} \\
\textit{Technidoo Solutions Germany }\\
Bavaria, Germany \\
klasen@technidoo.com}
}

\maketitle
\begin{abstract}
One of the successful early implementation of deep learning AI technology was on letter recognition. With the recent breakthrough of artificial intelligence (AI) brings more solid technology for complex problems like handwritten letter recognition and even automatic generation of them. In this research, we proposed deep learning framework called Ludwig AI Framework(LAIF) for Germany Suetterlin letter recognition and generation. To recognize Suetterlin letter, we proposed deep convolutional neural network. Since lack of big amount of data to train for the deep models and huge cost to label existing hard copy of handwritten letters, we also introduce the methodology with deep generative adversarial network to generate handwritten letters as synthetic data. Main source code is in ~\href{https://github.com/enkhtogtokh/LAIF}{https://github.com/enkhtogtokh/LAIF} repository. 
\end{abstract}


\section{Introduction}\label{intro}
The recent breakthrough of AI in many disparate fields is bringing us the more advanced technologies with industrial impacts. 
To make a step forward in this direction, we propose an efficient and flexible deep learning Ludwig AI (LAIF) framework, that have been conceived considering the main AI pipeline (from sensors to results) together with modern technology trends. The LAIF has two main pipelines which are deep convolutional neural network for the letter recognition and deep generative neural network for the script generation task. We applied them on Suetterlin script case. Generally it is possible to apply them on any similar type of tasks.

 The  Suetterlin was taught in all Germans from 1915 to 1941, the script is nowadays often used to refer to all varieties of old German handwriting. Suetterlin (in Germany, Sütterlin script, Sütterlinschrift) was the widely used form of Kurrent, the historical form of German handwriting that evolved alongside German blackletter typefaces. In German history, graphic artist Ludwig Sütterlin created a this handwriting script in 1911. 

The census shows in 1940, there was about 70 million population in Germany. However, note here, based on different sources shows up to 80 million population was in that time. Which means some tens of million people were writing the letters by their own handwritten style. It makes another challenging part for AI to recognize the vast amount of different handwritten letters recognition task. One possible solution here is to apply a generative AI techniques. Specially, the modern generative adverserial network (GAN)\cite{gan} generates variations of the letters. We propose the effective AI generative model in Section \ref{generation}. On other hand, modern state of the art deep learning models need big data, which possibly provides it. During Suetterlin script time, of course, all history and everything were left on it. The people, who are able to read and write on it, are getting old year by year. To decode mining resource information, it needs a person who could read the script and even to have solid understanding  in mining terminologies. There are inevitable demands to recognize the script with advanced technology. However, the research factually shows that we need to provide at least several modern researches, which initiated here as the first research work.

Concretely, the key contributions of the proposed work are:

\begin{itemize}
\item The first modern novelty feasibly AI technology for Suetterlin letter recognition
\item The generative AI technology for Suetterlin letter generation
\end{itemize}

Systematic experiments conducted on real-world acquired data have shown as: 
\begin{itemize}
\item It is possible to be common framework for any type of handwritten generation and recognition task.
\item It is possible to achieve 99.9\% accuracy on well prepared training data to recognize.
\item It is possible to generate realistic enough synthetic data generation with multiple variations
\end{itemize}

The rest of the paper is organized as follows.
The proposed framework is described in Section \ref{proposedarch}.
The recognition deep convolutional  model is explained in Section \ref{recognition}. 
The details about the implemented deep generative learning algorithms are discussed in Section \ref{generation} and the
experimental results are presented in Section \ref{experimentalresult}.
Finally, Section \ref{conclusion} provides the conclusions and future work.

\section{The Ludwig AI Framework (LAIF)}\label{proposedarch}
In this section, we discuss the proposed LAIF framework for Suetterlin letter AI applications.
The LAIF has two main pipelines which are the deep convolutional model recognition and deep generative model. We discuss them in detail with coming sections.

\subsection{The deep convolutional model to recognize}\label{recognition}
The transfer learning works in industry. It is computer vision task to recognize such visual letters. Which means convolutional type of deep neural network is the right model to tackle. We implemented the VGG19\cite{vgg19} model here. Optimizer algorithm and loss function are stochastic gradient descent and softmax cross entropy, respectively. For main hyper parameters, learning rate, weight decay, and momentum are 0.0001, 0.0001, and 0.9, respectively. Since Suetterlin has 30 alphabets, it has 30 new fine tuning classes. Specifically, we define the transfer learning model by\cite{enkhtogtokh}:
\begin{itemize}
\item prepare the pre-trained model
\item re-define the output layer as 30 neurons layer for the new task 
\item train the network 
\end{itemize}

This is called transfer learning, i.e. we have a model trained on another task, and we need to tune it for the new dataset we have in hand.    

\begin{algorithm}
\caption{Train} \label{alg:rec_train}
\begin{algorithmic}[1]
 
\STATE from mxnet import gluon 
\STATE from model\_zoo import get\_model
\STATE from mxnet.gluon import nn
\texttt{\\}
\texttt{\\}
\STATE  $train_data = gluon.data.DataLoader$
\STATE  $newclasses = 30$ 
\STATE  $finetunenet = get_model('vgg19', pretrained=True)$
\STATE  $finetunenet.output = nn.Dense(newclasses)$ 
\STATE  $finetunenet.output.initialize(init.Xavier(), ctx)$ 
\STATE  $finetunenet.collect_params().reset_ctx(ctx)$ 
\STATE  $metric = mx.metric.Accuracy()$ 
\STATE  $L = gluon.loss.SoftmaxCrossEntropyLoss()$
\FOR{epoch in range(epochs)} 
\STATE  $outputs = [finetunenet(X) for X in data]$
\STATE  $loss = [L(yhat, y) for yhat, y in zip(outputs, label)]$
\FOR{for l in loss} 
\STATE   l.backward()
\ENDFOR  
\ENDFOR   
\STATE $finetunenet.save_parameters('finetuned.params')$  
\end{algorithmic}
\end{algorithm}
 
In Algorithm \ref{alg:rec_train}, we use mxnet \cite{mxnet} with their latest deep learning framework GlounCV. As we described above, load convolutional model as VGG19, new output layer has 30 neurons, then iterations is looping for epochs on new dataset. Full source code is provided in appropriate repository.  

\subsection{The deep generative model to generate}\label{generation}
To generate Suetterlin script, we propose the deep convolutional generative adverserial neural network. Since it is generally computer vision problem, this variation of GAN is prepared to show more better results. Nature of GAN, in detail, it consists two adversarial networks as generative (G) and discriminative (D) models.  The generative model consists of assemble of convolutional layers and batch norms as shown in Algorithm \ref{alg:generative}. 

\begin{algorithm}
\caption{Generative model [G]} \label{alg:generative}
\begin{algorithmic}[1]
 
\STATE import torch
\STATE import torch.nn as nn
\texttt{\\}
\texttt{\\}
\STATE  $nn.BatchNorm2d(128)$
\STATE  nn.Upsample(scale\_factor=2)
\STATE  $nn.Conv2d(128, 128, 3, stride=1, padding=1)$ 
\STATE  $nn.BatchNorm2d(128, 0.8)$ 
\STATE  $nn.LeakyReLU(0.2, inplace=True)$ 
\STATE  $nn.Upsample(scale_factor=2)$
\STATE  $nn.Conv2d(128, 64, 3, stride=1, padding=1)$ 
\STATE  $nn.BatchNorm2d(64, 0.8)$ 
\STATE  $nn.LeakyReLU(0.2, inplace=True)$
\STATE  $nn.Conv2d(64, channels, 3, stride=1, padding=1),  nn.Tanh())$ 

\end{algorithmic}
\end{algorithm}

In Algorithm \ref{alg:convblock},  we define the convolutional block for Discriminator model.  The convolutional block consists of convlutional layer and batch norm. 

\begin{algorithm}
\caption{ConvBlock(input\_p, out\_p)} \label{alg:convblock}
\begin{algorithmic}[1]
 
\STATE import torch
\STATE import torch.nn as nn
\texttt{\\}
\texttt{\\}

\STATE  $convblock = []$
\STATE  $convblock.append(nn.Conv2d(input_p, out_p, 3, 2, 1)) $
\STATE  $convblock.append(nn.LeakyReLU(0.2, inplace=True))$
\STATE  $convblock.append(nn.Dropout2d(0.25))$
\STATE  $convblock.append(nn.BatchNorm2d(out_p, 0.8)))$ 
\STATE  return convblock
\end{algorithmic}
\end{algorithm}

The discriminator model [D] has four number of convolutional block as shown in Algorithm \ref{alg:discriminator}.

\begin{algorithm}
\caption{Discriminator model [D]} \label{alg:discriminator}
\begin{algorithmic}[1]
 
\STATE import torch
\STATE import torch.nn as nn
\texttt{\\}
\texttt{\\}

\STATE  $convblock(8, 16)$
\STATE  $convblock(16, 32)$
\STATE  $convblock(32, 64)$
\STATE  $convblock(64, 128)$
 
\end{algorithmic}
\end{algorithm}

For adversarial loss function, Binary Cross Entropy (BCE) implemented for the model. The Adam optimizer used for both [G] and [D] models. The Pytorch deep learning \cite{pytorch} framework is used for implementation.

\newpage
\section{Experimental Results}\label{experimentalresult}
In this section, we discuss first about the setup, and then evaluate the deep learning recognition and generation results are experimented in systematic scenarios. 
\subsection{Setup}
We train and test on ubuntu 18 machine with capacity of (CPU: Intel(R) Xeon(R) CPU @ 2.20GHz, RAM:16GB, GPU: NVidia GeForce GTX 1070, 16 GB).

\subsection{The recognition results}
Table \ref{tab:rec} shows the accuracy of training and validation on number of epochs. After 33 epochs, we achieved enough accuracy as loss, training, and validation are 100\%, 0,009, and 100\%, correspondingly. 

\begin{table} [h!]
\centering  
\begin{tabular}{||c c c c||} 
 \hline
 Number of epoch & Training accuracy & Loss & Validation accuracy \\  
 \hline\hline
 10 & 0.877 & 0.470 & 0.985 \\ 
 \hline
 30 & 0.985 & 0.025 & 1.000 \\
 \hline
 32 & 1.000 & 0.017 & 1.000 \\
 \hline
 33 & 1.000 & 0.009 & 1.000 \\
 \hline
\end{tabular}
\caption{The deep convolutional neural network recognition model training and validation accuracy on epochs.}
\label{tab:rec}
\end{table}

\begin{figure}[!htbp]
  \centering
  \includegraphics[width=\linewidth]{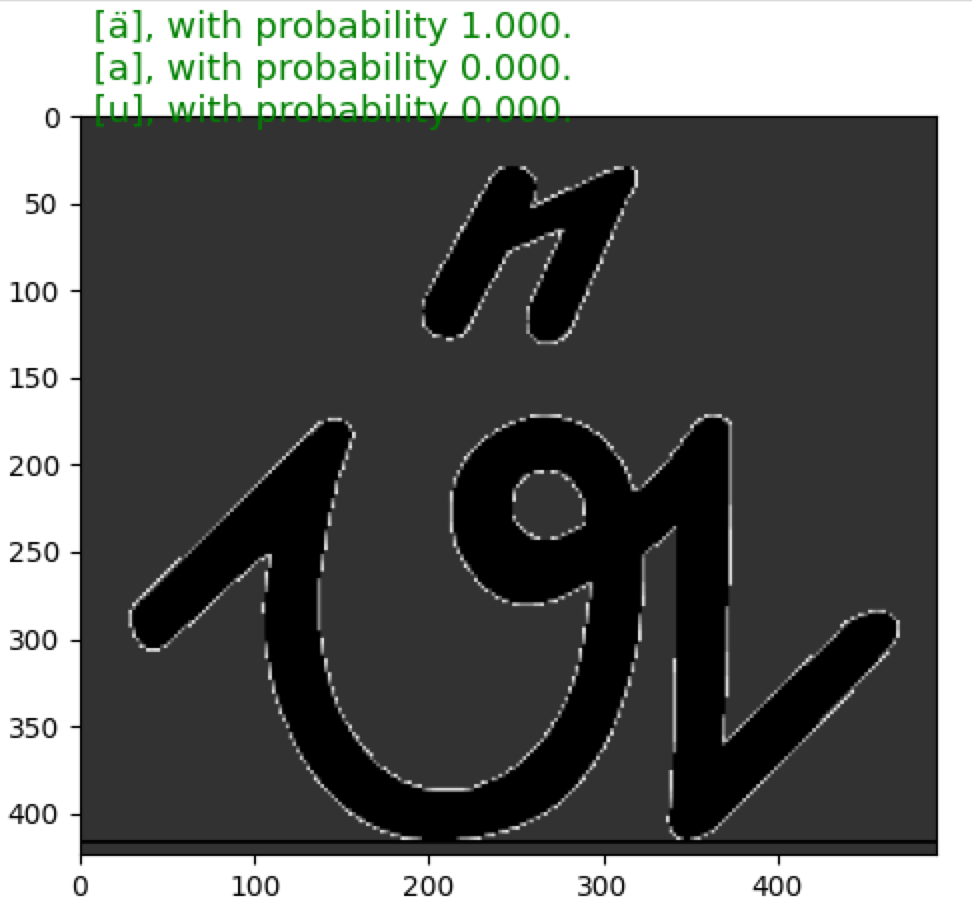}
  \caption{The recognition result for Suetterlin letter "ä"}
 
  \label{fig:rec_a}
\end{figure}

\begin{figure}[!htbp]
  \centering
  \includegraphics[width=\linewidth]{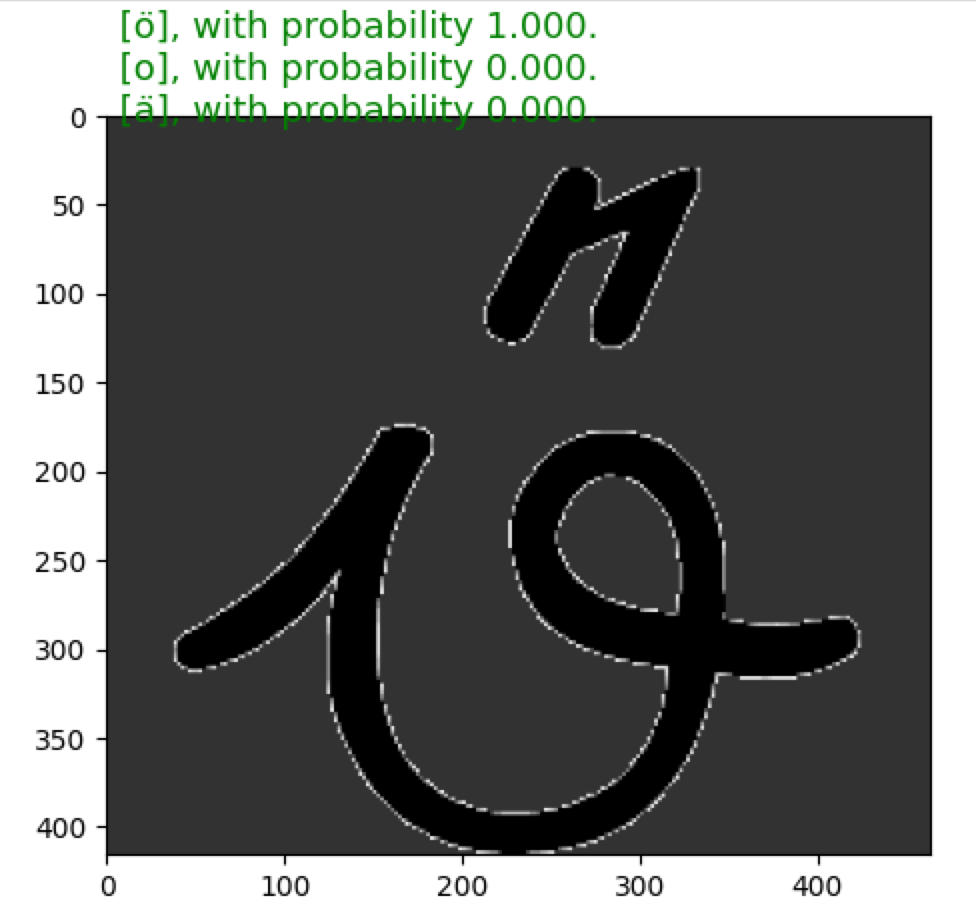}
  \caption{The recognition result for Suetterlin letter "ö"}
 
  \label{fig:rec_o}
\end{figure}

\begin{figure}[!htbp]
  \centering
  \includegraphics[width=\linewidth]{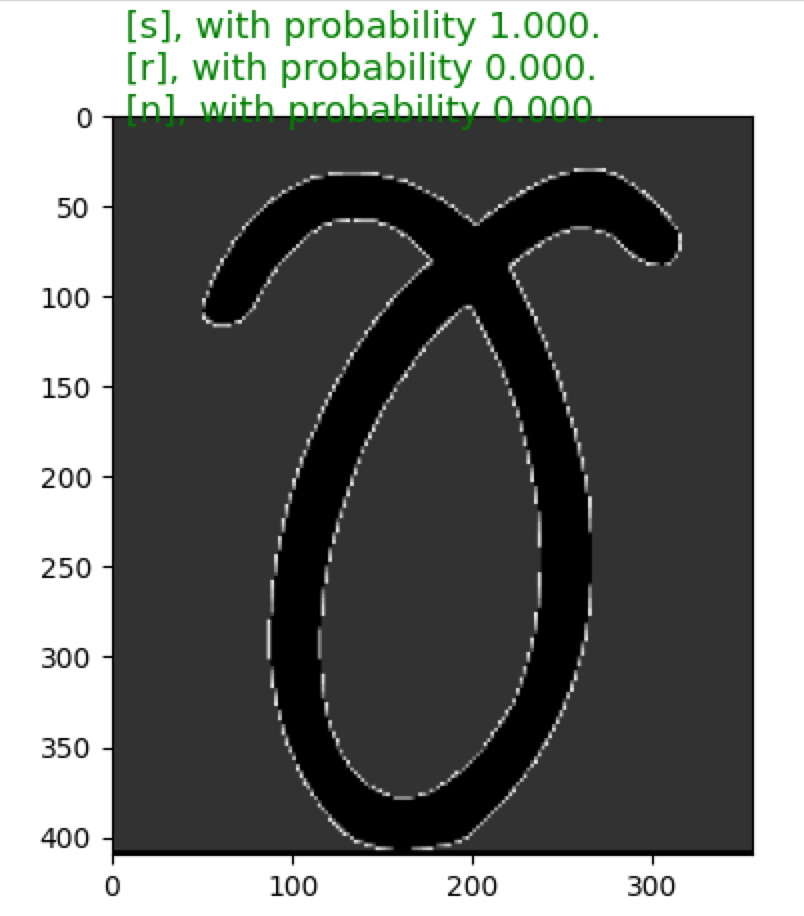}
  \caption{The recognition result for Suetterlin letter "s"}
 
  \label{fig:rec_s}
\end{figure}

Figure \ref{fig:rec_a}, \ref{fig:rec_o}, and \ref{fig:rec_s}  show the recognition results of ä, ö, and s Suetterlin letters, accordingly. We printed out top-3 probability classes. 
 
\subsection{The generation results}
\begin{figure}[!htbp]
  \centering
  \includegraphics[width=\linewidth]{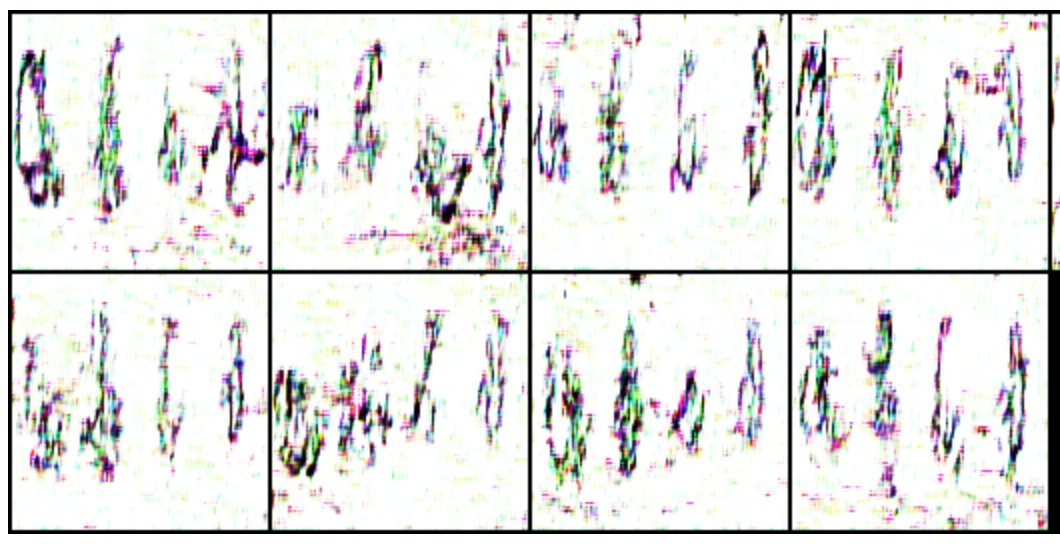}
  \caption{The generation result for Suetterlin script "ABCD" and "abcd" on 3000 epochs}
 
  \label{fig:gen_3000}
\end{figure}

\begin{figure}[!htbp]
  \centering
  \includegraphics[width=\linewidth]{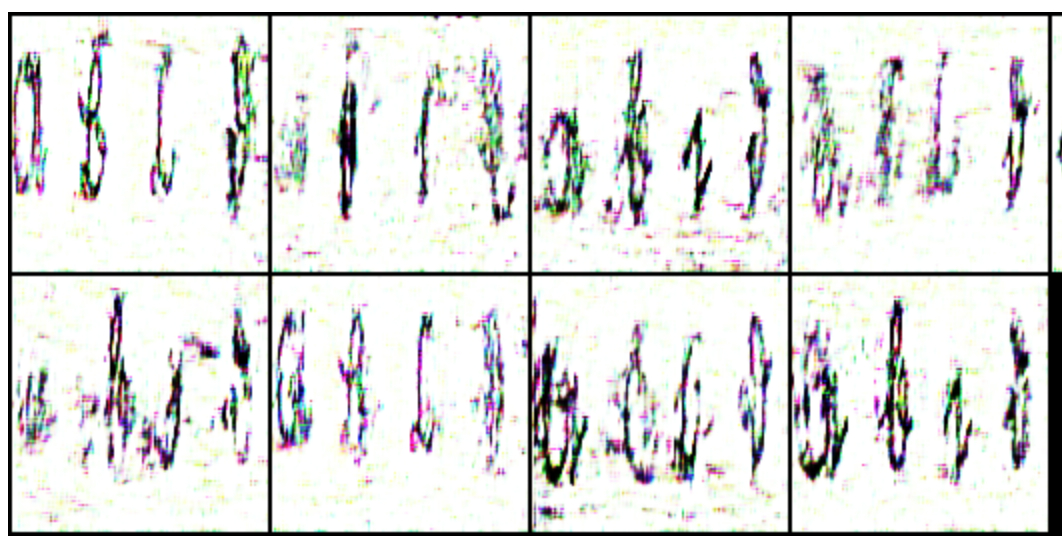}
  \caption{The generation result for Suetterlin script "ABCD" and "abcd" on 7000 epochs}
 
  \label{fig:gen_7000}
\end{figure}

\begin{figure}[!htbp]
  \centering
  \includegraphics[width=\linewidth]{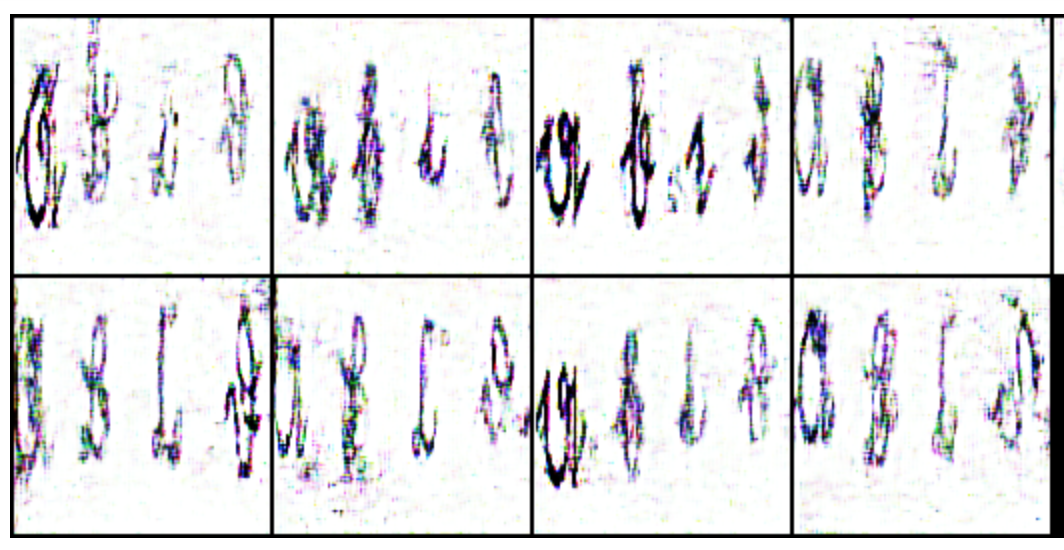}
  \caption{The generation result for Suetterlin script "ABCD" and "abcd" on 20000 epochs}
  \label{fig:gen_20000}
\end{figure}

In simple words, visual examination of results by humans is one of the most intuitive ways to evaluate generative neural network, since it is computer vision tasks.  In Figure \ref{fig:gen_3000}, \ref{fig:gen_7000}, and \ref{fig:gen_20000} show how generative deep adversarial neural network improves the generation epoch by epoch to generate Suetterlin simple "ABCD" and "abcd".

\begin{figure}[!htbp]
  \centering
  \includegraphics[width=\linewidth]{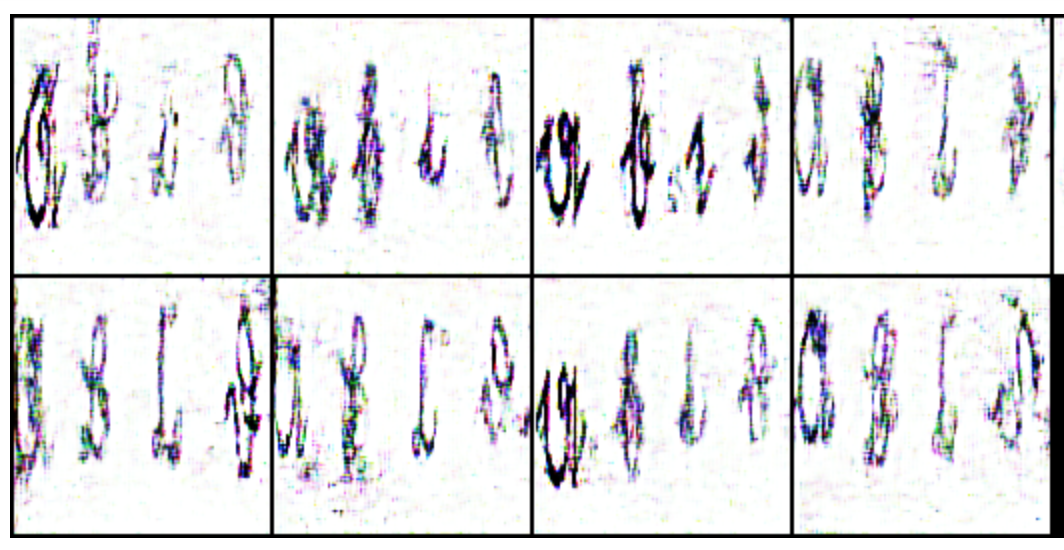}
  \caption{The generation result for Suetterlin script "ABCD" and "abcd" on 50000 epochs}
  \label{fig:gen_50000}
\end{figure}

Table \ref{tab:gen} shows the losses of discriminator and generator models on training epochs.

\begin{table} [h!]
\centering  
\begin{tabular}{||c c c||} 
 \hline
 Number of epoch & Discriminator loss & Generator Loss   \\  
 \hline\hline
 1080 & 0.072120 & 2.430481  \\ 
 \hline
 2900 & 0.015743 & 4.388611  \\
 \hline
 50000 & 0.012116 & 7.536002  \\
 \hline
\end{tabular}
\caption{The deep generative neural network recognition model training [D], [G] losses on training epochs.}
\label{tab:gen}
\end{table}

\newpage
\section{Conclusion}\label{conclusion}
We proposed the modern AI deep learning framework as LAIF for handwritten letters and application on Sueterllin script. Modern state-of-the-art deep learning approaches implemented to recognize the letters. And deep generative neural network proposed to generate automatically any possible variations of handwritten letters. All main important algorithms are directly provided in this research to develop first phase of Sueterllin script recognition and generation. The real visual results and some important evaluation accuracy scores are presented. 
In future works, we will publish next series of research to apply on special industrial sector of Suetterlin based on this architecture.

\vspace{12pt}


\newpage
\begin{thebibliography}{00}
\bibitem{enkhtogtokh}
Togootogtokh, Enkhtogtokh, and Amarzaya Amartuvshin. "Deep learning approach for very similar objects recognition application on chihuahua and muffin problem." arXiv preprint arXiv:1801.09573 (2018).

\bibitem{gan}
Goodfellow, Ian, et al. "Generative adversarial nets." Advances in neural information processing systems 27 (2014): 2672-2680.

\bibitem{mxnet}
Chen, Tianqi, et al. "Mxnet: A flexible and efficient machine learning library for heterogeneous distributed systems." arXiv preprint arXiv:1512.01274 (2015).

\bibitem{vgg19}
Carvalho, Tiago, et al. "Exposing computer generated images by eye’s region classification via transfer learning of VGG19 CNN." 2017 16th IEEE International Conference on Machine Learning and Applications (ICMLA). IEEE, 2017.
\bibitem{pytorch}
Paszke, Adam, et al. "Pytorch: An imperative style, high-performance deep learning library." Advances in neural information processing systems. 2019.
 


 
\end{thebibliography}
\end{document}